\documentclass[preprint,12pt]{elsarticle}

\usepackage{amssymb}
\usepackage[T1]{fontenc}
\usepackage{microtype}
\usepackage{graphicx}
\usepackage{amsmath}
\usepackage{amsthm}
\usepackage{booktabs}
\usepackage{algorithm}
\usepackage{algorithmic}
\usepackage{latexsym}
\usepackage{amssymb}
\usepackage{multirow}
\usepackage{microtype}
\usepackage{float}
\usepackage{booktabs}
\usepackage{times}
\usepackage{soul}
\usepackage{url}
\usepackage{color}
\usepackage{makecell}
\usepackage{stfloats}

\journal{NeuroComputing}

\begin{document}

\begin{frontmatter}



\title{Learning a General Clause-to-Clause Relationships for Enhancing Emotion-Cause Pair Extraction}



\author{Hang Chen, Xinyu Yang, Xiang Li \\
Xi'an Jiaotong University, No.28, Xianning West Road, Xi'an, Shaanxi, 710049, P.R. China\\
  Du Xiao Man Inc., No.10, Xibeiwang West Road, Beijing, China \\
  \texttt{albert2123@stu.xjtu.edu.cn, yxyphd@mail.xjtu.edu.cn, lixiang2work@163.com} \\
}

\begin{abstract}
    Emotion-cause pair extraction (ECPE) is an emerging task 
    aiming to extract potential pairs of 
    emotions and corresponding causes from documents. 
    Previous approaches have focused on modeling 
    the pair-to-pair relationship and achieved promising results.  
    However, the clause-to-clause relationship, which fundamentally 
    symbolizes the underlying structure of a document, 
    has still been in its research infancy. In this paper, 
    we define a novel clause-to-clause relationship. 
    To learn it applicably, 
    we propose a general clause-level encoding model named \textbf{EA}-GAT 
    comprising \textbf{E}-GAT and \textbf{A}ctivation Sort.   
    E-GAT is designed to aggregate information from 
    different types of clauses;  
    Activation Sort leverages the individual emotion/cause 
    prediction and the sort-based mapping to propel the 
    clause to a more favorable representation. 
    Since EA-GAT is a clause-level encoding model, 
    it can be broadly integrated with any previous approach. 
    Experimental results show that our approach 
    has a significant advantage over all current approaches 
    on the Chinese and English benchmark corpus, 
    with an average of $2.1\%$ and $1.03\%$. 

\end{abstract}




\end{frontmatter}



\section{Introduction}

Recently, fine-grained understanding of emotional clause level 
~\citep{turcan-etal-2021-multi,poria2020beneath,poria2021recognizing} 
has gained increasing attention from the research community, 
especially a surge of research to determine 
the cause of emotional expression in text
~\citep{bi2020ecsp,ding2020experimental,li2022neutral,uymaz2022vector}. 

Emotion-Cause Pair Extraction (ECPE)
~\citep{xia-ding-2019-emotion}
is a new task that extracts all emotion clauses coupled with their 
cause clauses from a given document. 
For understanding simplicity, we following
~\citet{wei-etal-2020-effective,chen-etal-2020-conditional} 
abstract the process of ECPE as four separate modules:  
word-level encoding, clause-level encoding, 
pair-level encoding, and pair extraction. The first three 
parts separately model the word-to-word relationship, 
clause-to-clause relationship, and pair-to-pair relationship. 

\begin{figure}[hbp]
  \centerline{\includegraphics[width=0.5\linewidth]{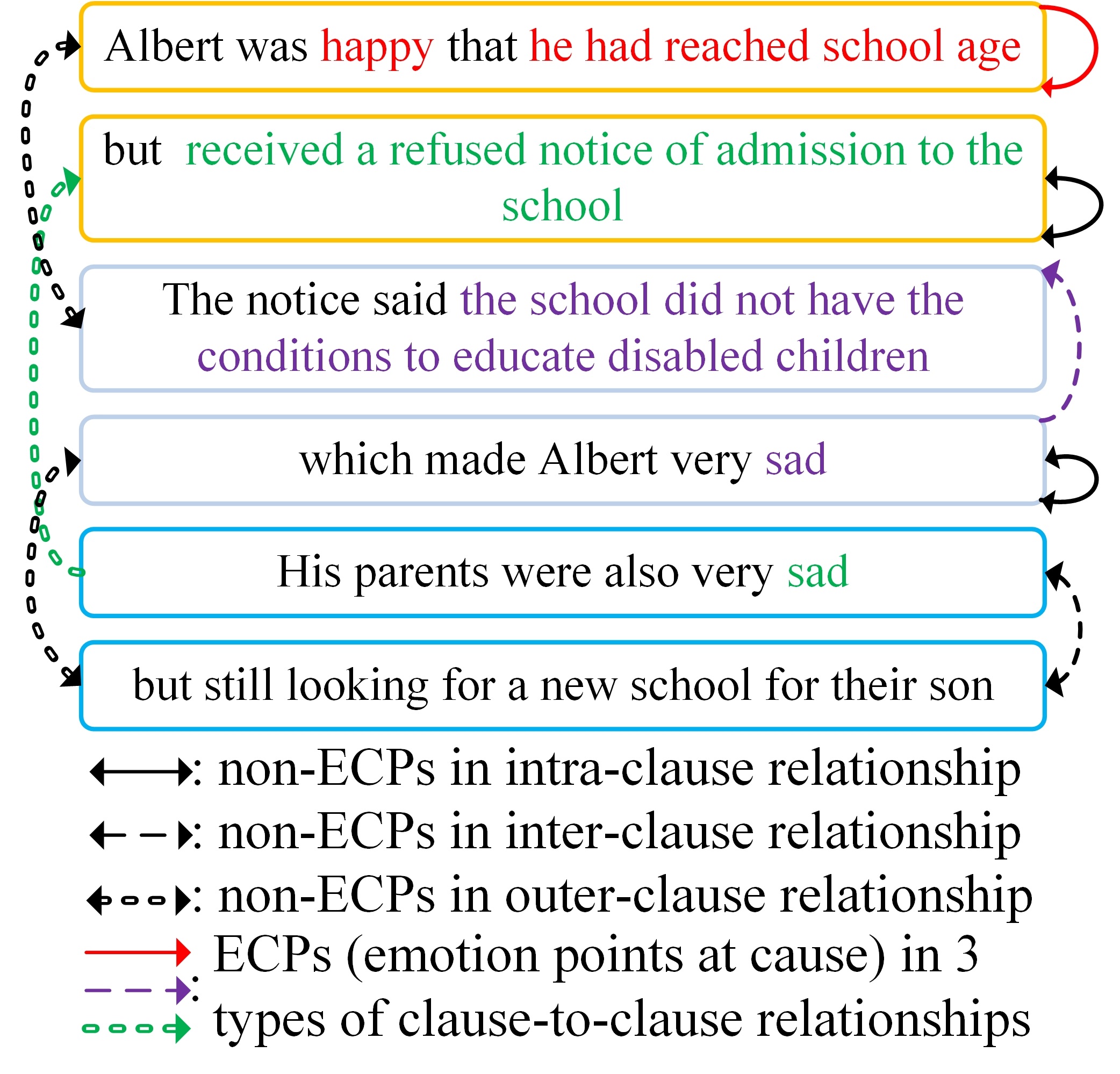}}
  \caption{A sample document with three sentences. Each sentence 
  consists of two clauses. There are three types of 
  ECPs inked red, purple, and green.}
  \label{figtask}
\end{figure}

For the pair-to-pair relationship, there have been numerous works
~\citep{fan-etal-2020-transition,ding-etal-2020-ecpe,fan2022combining,qie2022dcn,jin2022emotion} 
to contribute particular advantages with their different networks.  
However, for the clause-to-clause relationship, 
most approaches use BiLSTM and only~\citet{wei-etal-2020-effective} 
adopts GAT~\citep{velivckovic2017graph}. 
From~\citet{thost2021directed,shen2021directed}, 
BiLSTM relatively neglects the remote clause, and GAT loses 
the sequential information. As a result, clause-level encoding 
is a fundamental yet under-explored part of the ECPE process. 
And this imbalance is intelligible because pair-level encoding
has an exclusive direct influence on the pair extraction.

Moreover, from the findings of linguistics and psychologic: 
~\citet{mann1988rhetorical,marcu2000theory}  
have proposed that clauses or sentences are often indicative of an 
underlying structure of one document;   
~\citet{ruusuvuori2012emotion} has provided support that  
the emotion of clauses tends to be constant in a sentence. 
More recently, ~\citet{wu2022knowledge} has argued that the entire event 
is usually across a set of clauses in a sentence. 
~\citet{huang2022deep} has exploited general grammatical conventions 
to span-encode sentences and clauses, demonstrating the existing 
grammatical information contained in sentences and clauses. 
Hence, advances in the clause-to-clause relationship, 
specifically the knowledge of whether the clause 
spans a sentence, are available for 
distinguishing Emotion-Cause Pair (ECP) from the other non-ECPs. 

To this end, we define a novel   
clause-to-clause relationship, as shown in Figure\textcolor{red}{\ref{figtask}}. 
Proceeding from the sentence attributions of two observed clauses, 
we delimit clause-to-clause relationships as:
\begin{itemize}
  \item \textbf{Outer-clause relationship}: The two observed 
  clauses are in different sentences. 
  \item \textbf{Inter-clause relationship}: The two different observed
  clauses are in the same sentences. 
  \item \textbf{Intra-clause relationship}: The two observed 
  clauses are the same clause (self-influence). 
\end{itemize}

Furthermore, \citet{chen2022recurrent,shi2022optimizing} argued 
that the clause representations can be integrated with 
the prediction sequence of two subtasks named emotion extraction 
and cause extraction to achieve improvement. 
Nevertheless, due to the fine-grained relationship representation 
defined by ours 
mismatching the binary-classification prediction sequence, 
such the straightforward integrating hides the clues to emotions.  
\citet{chen-etal-2020-unified} has reformulated binary classifiers 
as 7-classifier and 4-classifier to fine the grain but 
at the loss of generality. 

Considering this problem, 
and effectively empowering all approaches to utilize 
our defined relationship's knowledge, we propose a clause-level 
encoding model for ECPE, instead of contributing an end-to-end model. 
Our model, composed 
of \textbf{E}nhanced GAT and \textbf{A}ctivation Sort, is called EA-GAT, 
which can be easily applicable to all existing approaches. 
Enhanced GAT is a multi-mask GAT proposed to learn our new defined 
clause-to-clause relationship. The Activation Sort is  
a radical but valid pre-processing trick for achieving grain consistency. 

Specifically, Enhanced GAT 
yields a clause representation through separately 
modeling the attention of 3 types of relationships. 
And the Activation Sort is fed with emotion/cause prediction sequences 
and outputs the auxiliary information for the second Enhanced GAT module. 
It wides the gap among predictions via sort-based mapping, which 
can explore different relationships of clauses beyond the polarity prediction.  
Finally, EA-GAT produces a more favorable clause representation 
for subsequent pair-level encoding.   

To gauge the generality of our clause encoding model, we conduct 
comprehensive experiments comparing with all published ECPE methods. 
And we perform intrinsic contrast experiments of the adjacency matrix to 
obtain insight into how EAGAT grasps the clause relationship. 
Furthermore, we construct a new benchmark of the English dataset via 
leveraging the RECCON proposed for Causal Emotion Entailment task. 

Our main contributions can be summarized as follows: 
\begin{itemize}
  \item To the best of our knowledge, we are first to 
  define the clause-to-clause relationship and develop 
  a multi-mask attention to learn their representation.  
  \item We proposed the EA-GAT model, which can more effectively work 
  in the clause-level encoding part of ECPE than networks adopted by previous methods. 
  \item We conduct our comprehensive evaluation on the standard Chinese dataset 
  and a newly introduced English dataset, 
  which demonstrates that our method can be generally embedded into 
  all existing approaches and significantly outperforms 
  their original performance, including the state-of-the-art work.
\end{itemize}

The remainder of this paper is organized as follows: 
in Section~\ref{rw}, we briefly introduce the related works; 
in Section~\ref{td}, we formalize the task; 
in Section~\ref{me}, we describe our proposed method; 
in Section~\ref{ex}, we present the details and data of the experiments; 
in Section~\ref{rd}, we analyze the results and limitations; 
in Section~\ref{co}, we conclude of the proposed work.

\section{Related Works}\label{rw}

\begin{table*}
    \footnotesize
    \centering
    \resizebox{\linewidth}{!}{
    \begin{tabular}{|l|c|c|c|c|}
      \hline
      \textbf{Model} & word-level encoding & clause-level encoding & pair-level encoding&clause extractions\\
      \hline
      Inter-EC \citep{xia-ding-2019-emotion} & word2vec       & BiLSTM & Logistic regression&\multirow{17}{*}{\thead{linear layer\\+softmax}} \\
      RankCP \citep{wei-etal-2020-effective}   & word2vec/BERT  & GAT              & Rank model&\\ 
      ECPE-2D \citep{ding-etal-2020-ecpe}  & word2vec/BERT  & BiLSTM & transformer&\\
      TDGC \citep{fan-etal-2020-transition}     & word2vec/BERT  & BiLSTM & Transition Model&\\
      IE-CNN \citep{chen-etal-2020-unified}   & word2vec       & BiLSTM & CNN&\\
      PairGCN \citep{chen-etal-2020-end}  & word2vec/BERT       & BiLSTM& GCN&\\
      SLSN \citep{cheng-etal-2020-symmetric}     & word2vec		  & BiLSTM & Local Searcher&\\
      JointNN \citep{tang2020joint}  & BERT    		  & BiLSTM+attention & Joint MLP Model&\\
      MLL \citep{ding-etal-2020-end} & word2vec       & BiLSTM & sliding window&\\
      CPAM \citep{chen-etal-2020-conditional}	 & word2vec       & BiLSTM & Aggreggation&\\  
    RSN \citep{chen2022recurrent} & word2vec/BERT & BiLSTM &Linear&\\
    MASTM \citep{shi2022optimizing} & word2vec &BiLSTM &3-level filters&\\
    PTF \citep{wu2023pairwise} & word2vec/BERT & BiLSTM & PTF-based model & \\
    CL-ECPE \citep{zhang2022cl} & word2vec & BiLSTM & adversarial model &\\
    BERT+ \citep{cao2022research} & BERT & BiLSTM & MOO &\\
    MGSAG \citep{bao2022multi}&word2vec/BERT & BiLSTM& Semantic Aware Graph &\\
    SAP-ECPE \citep{huang2022deep}&word2vec&BiLSTM&BiGRU&\\
    \hline
    \end{tabular}}
    \caption{ 4 parts of existing approaches. There are different networks for encoding the pair-to-pair representations whereas clause-level encoding sharply lacks effective research. }
    \label{tabrelated}
  \end{table*}

The process of ECPE task has been carried out in $3$ encoding parts  
and $1$ extraction part, as shown in Table\ref{tabrelated}. 
In word-level encoding, word representation adopts 
pre-training with word2vec \citep{NIPS2013_9aa42b31} toolkit or  
$BERT$~\citep{devlin-etal-2019-bert}. 
Moreover, almost all approaches adopt BiLSTM to model the relationship 
between clauses in clause-level encoding. 
But pair-level encoding has attracted 
substantial interest from numerous researchers, in which 
\citet{fan-etal-2020-transition} has transformed ECPE into a procedure of directed graph structure and adopted a transition-based model to identify the semantic relationship between emotions and causes; 
\citet{wei-etal-2020-effective} has used the RankNN with an RBF kernel function for the pair-level encoding. Specifically, they proposed a GAT to capture the underlying relationship among different clauses via designing a full-connected graph;  
\citet{chen-etal-2020-end} has adopted GNN framing pairs as nodes to learn the pair-level representation. 

And to alleviate the loss of information by binary classification, 
\citet{chen-etal-2020-unified} used a more fine-grained tagging scheme that combines emotion labels and cause labels with emotion types separately, and 
\citet{chen-etal-2020-conditional} has defined a new task of determining whether or not an ECP has a valid causal relationship under different contexts. 
\citet{wu2023pairwise} propose a labeling framework tackling the whole 
ECPE in one unified tagging task to optimize the ECPE task globally and extract more accurate ECPs. 
Based on these attempts, increasing works have been made to 
design the extra annotation~\citep{chang2022emotion,huang2022deep} or auxiliary dataset~\citep{cao2022research,wu2022knowledge,zhang2022cl} or keywords~\citep{bao2022multi}.

More recently, the clause-level relationship between emotions and causes has gained 
significant attention due to its effectiveness in real-world applications.
~\citet{kumar2022emotion} has replaced the traditional technique 
for rule-based semantic analysis work on sentence-level 
with the ECPE-BERT model to recognize emotion in psychological texts. 
And~\citet{ghosh2022cares} has proposed a pre-trained transformer-based 
ECPE model to address the problem of cause annotation and cause extraction for emotion in suicide notes datasets. Furthermore, inspired 
by recent advances in ECPE task,~\citet{halat2022multi} has analyzed the 
causes leading to a negative statement via ECPE model, to improve the 
recognition of hate speech and offensive language (HOF). 

Of 4 parts of the ECPE process, our model, as a clause-level encoding method, is broadly applicable 
to all existing works well, reserving 
the crucial components of each work 
and exploiting the clause-to-clause relationships.

\section{Task Definition}\label{td}

Given a document $D=(c_{1},c_{2},\dots,c_{|D\vert })$ 
where $D$ is the number of 
clauses and the $i$-th clause 
$c_{i}=(\omega^{i}_{1},\omega^{i}_{2},\dots,\omega^{i}_{|c_{i}\vert })$ 
is a word sequence, 
the goal of ECPE is to extract 
a set of emotion-cause pairs (ECPs) in $D$:

\begin{equation}
  P=\left\{\dots,(c_{i}^{e},c_{j}^{c}),\dots \right\}
  \label{eqn1}
\end{equation}

where $(c_{i}^{e},c_{j}^{c})$ is an ECP consisting of 
$i$-th clause $c_{i}^{e}\in D$ as emotion clause 
and $j$-th clause $c_{j}^{c}\in D$ as corresponding cause clause. 
Additionally, the input of our method is 
$(c_{1},c_{2},\dots,c_{|D\vert })$ outputting 
from word-level context encoding, 
and the output of our method is clause representation 
(${h^{'}_{c_{1}},h^{'}_{c_{2}},\dots,h^{'}_{c_{|D|}}}$). 
Note that the pair representation $P_{i,j}$, the input 
of pair-to-pair encoding, is contacted by 
clause representation $h^{'}_{c_{i}}$ and $h^{'}_{c_{j}}$, 
which is entirely different from clause representation and 
not developed in this paper. 

\section{Methodology}\label{me}

\begin{figure*}[hbp]
    \includegraphics[width=0.95\linewidth]{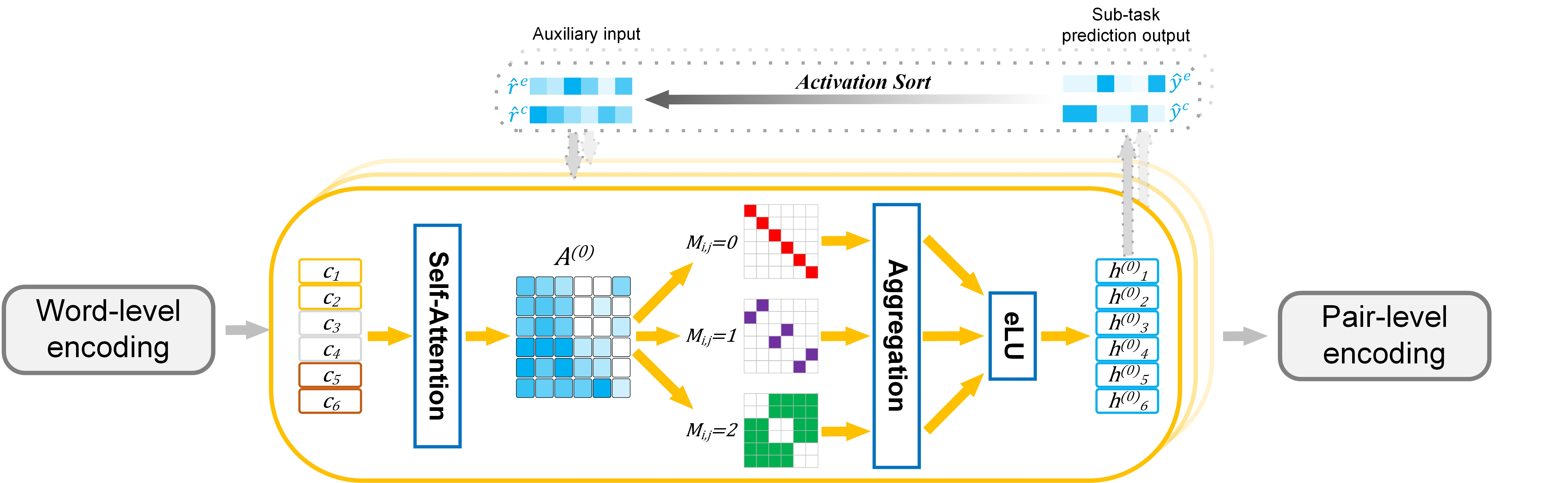}
    \caption{Overview of our model with the orange-dash box 
    indicates our model, and the gray box represents 
    our model-irrelevant parts in ECPE processing. 
    The input of our model is individual clause 
    embedding integrated by word-level encoding and the output of 
    our model is the optimal clause representation for  
    pair-level encoding. }
    \label{figoverall}
  \end{figure*}

The main framework of EA-GAT consists of hierarchical \textbf{E}nhanced 
GAT modules and \textbf{A}ctivation Sort modules. 
The overview is shown in Figure\ref{figoverall}. Enhanced GAT is used to 
process clauses under the clause-to-clause relationship 
defined in Section\ref{sec4.1} and Activation Sort enlarges the 
spread of predictions weighting of negative samples to form the 
new adjacency matrix for Enhanced GAT. 
Furthermore, an additional sort loss 
is adopted to evaluate the performance of the whole framework. 

\subsection{Building Clause-to-Clause Relationship}\label{sec4.1}

In the tokenizer, 3 sentence-ending dot marks, ``$.$'',``$?$'' and``$!$'', 
are denoted by a token <$period$>. 
For notational consistency, we also add <$period$> to the beginning 
and the end of a document. 
Hence, for any sentence $S$, one can partly define a unique sequence order 
on a document $D=(<period>,c_{1},c_{2},\ldots,<period>,c_{i},\ldots,c_{j},<period>,\ldots,c_{|D|},<period>)$, 
such that all clauses between two contiguous <$period$> belong to 
the same $S$. 
Then, a document 
can also be denoted by $D=(S_{1},S_{2},……,S_{n})$,  
where $n$ is one less the number of <$period$>. 

We formalize the clause-to-clause relationship by the 
multi-mask matrix 
$M=\in \mathbb{R}^{|D|\times |D|}$ as follows: 

\begin{equation}
  \begin{split}
    M_{i,j}=\left\{
      \begin{array}{lr}
        2, &S_{i}\neq S_{j}\\
        1, &S_{i}=S_{j}  \bigwedge  c_{i}\neq c_{j}\\
        0, &c_{i}=c_{j}
      \end{array}
    \right.
  \end{split}
  \label{eqn2}
\end{equation}

where $M_{i,j}$ denotes the relationship type from $c_{j}$ to $c_{i}$. 
There are 3 diverse values $(2,1,0)$ that separately represent 
outer-clause relationship, inter-clause relationship,  
and intra-clause relationship. 

\subsection{ Enhanced Graph Attention}

Given a clause embedding $D=(c_{1},c_{2},\dots,c_{|D\vert })$ 
generated by word-level encoding, 
we design Enhanced GAT to instantiate the 
information aggregation of clauses. 

Specifically, for each clause, self-attention  aggregates 
information from neighboring clauses to learn the 
updated attention weight as follows:

\begin{equation}
  A^{(0)}_{ij}=\frac{LeakyReLU(e_{i,j}^{(0)})}{\sum_{k\in|D|}LeakyReLU(e_{i,k}^{(0)})} 
\label{eqn3}
\end{equation}

\begin{equation}
  e_{i,j}^{(0)}=W^{(0)}_{i(row)}\overrightarrow{c}_{i}+W^{(0)}_{j(col)}\overrightarrow{c}_{j} 
  \label{eqn4}
\end{equation}

Where $A^{(0)}$represents the adjacency matrix modeled 
by graph attention, the superscript $(0)$ denotes 
the first E-GAT module ($(1)$ denotes the second),  
and $A^{(0)}_{ij}$ measures
the importance of influence of the $j$-th clause 
on the $i$-th clause, 
$W^{(0)}_{row}$ and $W^{(0)}_{col}$ are the 
learnable parameters in the graph attention. 

Considering the multi-mask $M$, the primary clause representation 
$h$ is the sum of 3 types of relationship attention operators: 

\begin{equation}
  h=\sum_{m=0,1,2} eLU(A_{m}^{(0)}DW^{(0)}_{m})
  \label{eqn5}
\end{equation}

\begin{equation}
  \begin{split}
    A^{(0)}_{m(i,j)}=\left\{
      \begin{array}{lr}
        A^{(0)}_{i,j}, &M_{i,j}=m\\
        0, &M_{i,j}\neq m
      \end{array}
    \right.
  \end{split}
  \label{eqn6}
\end{equation}

Where the  $W^{(0)}_{m}$ denotes the learnable parameters for 
each mask value $m$. Following the strategy of other ECPE approaches 
which performs two individual subtasks named emotion/cause extractions, 
the probabilities of each clause for the two subtasks are computed as follows:

\begin{equation}
  \widehat{y}^{e}=Sigmoid(W_{e}h+b_{e})
  \label{eqn7}
\end{equation}

\begin{equation}
  \widehat{y}^{c}=Sigmoid(W_{c}h+b_{c})
  \label{eqn8}
\end{equation}

Where $\widehat{y}^{e}$ and $\widehat{y}^{c}$ are 
prediction sequences of emotion/cause extraction, 
$W_{e}$, $W_{c}$, $b_{e}$ and $b_{c}$ are learnable parameters 
of two linear layers.

\subsection{Activation Sort}\label{subas}

We use prediction sequences 
$\widehat{y}^{e}$ and $\widehat{y}^{c}$ 
to generate an enhanced adjacency matrix. 
Because as an individual subtask, 
$\widehat{y}^{e}_{i}$ 
denotes the probability that the $i$-th clause 
is an emotion clause. Similarly, 
$\widehat{y}^{c}_{j}$ 
denotes the probability that the $j$-th clause 
is the cause of a certain emotion clause.  
For an ECP $(c_{i}^{e},c_{j}^{c})$, 
it should more possibly consist of the $c_{i}^{e}$ 
with a higher $\widehat{y}^{e}_{i}$ and the $c_{j}^{c}$ 
with a higher $\widehat{y}^{c}_{j}$. 

But the ground truth sequences $y^{e}$ and $y^{c}$ are binary,
~\citet{chen-etal-2020-unified,chen-etal-2020-conditional} have 
provided the supports of which 
framing emotion/cause extractions as binary classification 
partly ignores the relationship from cause to a specific emotion. 

In order to make emotion/cause predictions more efficient 
to indicate the relationship among all clauses, especially non-ECPs. 
We design a trick processing the emotion/cause predictions to provide 
EGAT auxiliary information as part of input. 
In the output of each EGAT module, We sort all predictions 
in $\widehat{y}^{e}$ and $\widehat{y}^{c}$ to expand the 
differences, which contributes to the next EGAT to generate more favorable adjacency. 
Our sort approach named $Activation Sort$ is 
operated as follows: 

\begin{equation}
  \begin{aligned}
  \widehat{r}^{e}_{i}=label[argsort(\widehat{y}^{e})]
  \end{aligned}
  \label{eqn9}
\end{equation}

\begin{equation}
  \begin{aligned}
  \widehat{r}^{c}_{j}=label[argsort(\widehat{y}^{c})]
  \end{aligned}
  \label{eqn10}
\end{equation}

\begin{equation}
  \begin{aligned}
  label=[1,2,3,\dots,|D|]
  \end{aligned}
  \label{eqn11}
\end{equation}

Specifically, 
we decreasingly sort the $\widehat{y}^{e}$ and $\widehat{y}^{c}$, 
and separately give each 
$\widehat{y}^{e}_{i}$ and $\widehat{y}^{c}_{j}$ 
a serial number $\in [1,|D|]$ 
according to the sorting position, 
use this serial number as an index, 
and make the value of the corresponding index in the $label$ 
assigned to $\widehat{y}^{e}$ and $\widehat{y}^{c}$.  
We design the $label$ as a sequence of natural numbers 
not containing $0$, 
which can distinguish the difference between each prediction. 
A sort position embedding is adopted 
for $\widehat{r}^{e}$ and $\widehat{r}^{c}$ 
to regulate magnitude order.  
And after experimental trying, We also defined 
an alternative $label$ sequence as follows:

\begin{equation}
  \begin{aligned}
  label[i]=2^{i}(i=[\log^{|D|}_{2}]-[\log^{|D|-i+1}_{2}])
  \end{aligned}
  \label{eqn12}
\end{equation}

We exemplify this equation by a document with $10$ clauses, 
the value of $label$ is $[1,1,1,2,2,2,2,4,4,8]$. 
In this way,  $\widehat{r}^{e}$ and $\widehat{r}^{c}$ 
have more visible differences compared with 
the original prediction sequence close to $[0,0,0,0,0,0,0,0,1,1]$ (two positive example clauses). 

Note that: (i) there are countless  $label$ sequences  
that can expand the difference not be put forward; 
(ii) compared with popular activation functions 
such as $sigmoid$ or $tanh$, $Activation Sort$ can not 
only expand the distance among non-ECPs after normalization 
but also narrow down the distance between ECPs and non-ECPs; 
(iii) the gradient stopping caused by argsort is a benefit 
to eliminate the negative impact of 
$Loss_{non-ECP}$ on $P^{0}_{y^{pair}}$ 
(detail shown in Section\ref{sec4.5}); 
(iv) in the pre-output of subtasks, $\widehat{y}^{e}$ and $\widehat{y}^{c}$ 
reserve the original prediction sequences.

\subsection{Enhanced GAT with Activation Sort}

After sorting $\widehat{y}^{e}$ and $\widehat{y}^{c}$, 
activated clause-level relationship 
(adjacency) matrix $e^{(1)}_{i,j}$is 
denoted as: 

\begin{equation}
  e^{(1)}_{i,j}=W^{(1)}_{i(row)}\widehat{r}^{e}_{i}+W^{(1)}_{j(col)}\widehat{r}^{c}_{j}
  \label{eqn13}
\end{equation}

\begin{equation}
  A^{(1)}_{ij}=\frac{LeakyReLU(e_{i,j}^{(1)})}{\sum_{k\in|D|}LeakyReLU(e_{i,k}^{(1)})} +A^{(0)}_{i,j}
\label{eqn14}
\end{equation}

Where $W^{(1)}_{i(row)}$ and $W^{(1)}_{j(col)}$ 
are learnable parameters that separately represent 
the correlation of clauses in $\widehat{r}^{e}_{i}$ 
and $\widehat{r}^{c}_{j}$. 
Under a hypothesis that $\widehat{y}^{e}$ and $\widehat{y}^{c}$ 
are close to ground truth sequences $y^{e}$ and $y^{c}$ (from 
the empirical result of Section\ref{sec5.3}, it is a relatively weak 
hypothesis),  $A^{(1)}$ depicts a learnable clause-to-clause 
relationship adjusted by the emotion/cause ground truth.

Then we feed $A^{(1)}$ into the Enhanced GAT again, aiming to 
generate the clause representation $h^{(1)}$ as follows:

\begin{equation}
  h^{(1)}=\sum_{m=0,1,2} eLU(A_{m}^{(1)}DW^{(1)}_{m})
  \label{eqn15}
\end{equation}

where $h^{(1)}$ is the output as enhanced clause representation 
that can be used in the follow-up pair-level encoding 
and extraction. The empirical case in Section\ref{sec5.5} 
holds the interpretability of $h^{(1)}$ and $h$. On the whole, 
Starting with the $2$-th EGAT, each adjacency matrix will be added the 
auxiliary information from the output of the previous EGAT. We denote 
the layers of EGAT as $l=2,3,\dots,L$, the clause representation of 
each EGAT can be writen as: 

\begin{equation}
  h^{(l)}=\sum_{m=0,1,2} eLU(A_{m}^{(l)}DW^{(l)}_{m})
  \label{eqn15}
\end{equation}

\begin{equation}
  A^{(l)}_{ij}=\frac{LeakyReLU(e_{i,j}^{(l)})}{\sum_{k\in|D|}LeakyReLU(e_{i,k}^{(l)})} +A^{(l-1)}_{i,j}
\label{eqn14}
\end{equation}

\subsection{Optimization}\label{sec4.5}

\begin{figure*}[hbp]
    \includegraphics[width=\linewidth]{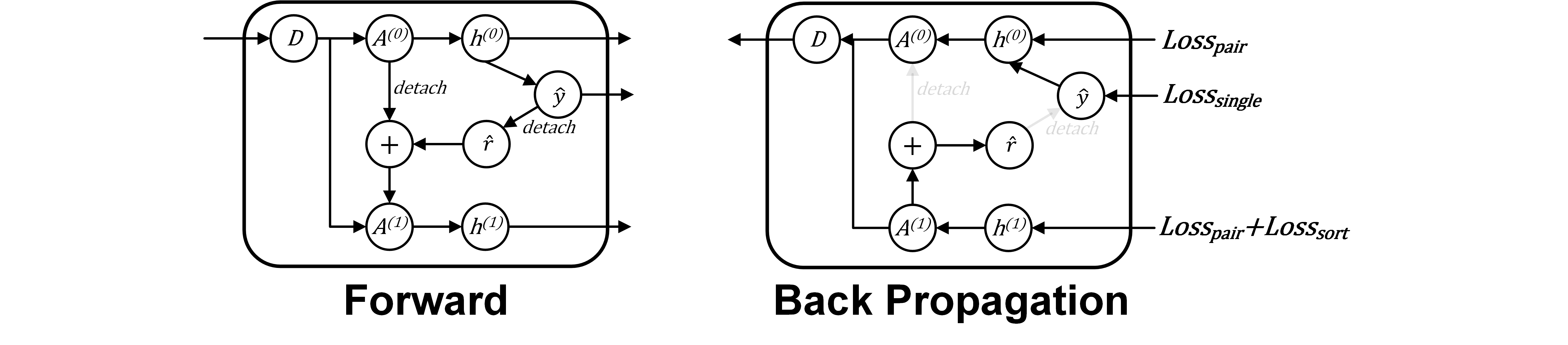}
    \caption{The forward processing and back propagation processing 
    of our model (We only show the first and second layers of EGAT). }
    \label{figsort}
  \end{figure*}

Generally, ECPE task is optimized by 3 cross-entropy errors: 
$Loss_{e}$, $Loss_{c}$ and $Loss_{pair}$, 
where $Loss_{e}$ and $Loss_{c}$ measure 
 $\widehat{y}^{e}$ and  $\widehat{y}^{c}$.  $Loss_{pair}$ 
measures the output of pair prediction sequence. 

\begin{equation}
  Loss_{e}=\sum_{l = 1}^{L}\sum_{i \in |D|} -(P^{l}_{y^{e}_{i}} \log(1-P^{l}_{\widehat{y}^{e}_{i}})+(1-P^{l}_{y^{e}_{i}}) \log P^{l}_{\widehat{y}^{e}_{i}}  )
  \label{eqn16}
\end{equation}

\begin{equation}
  Loss_{c}=\sum_{l = 1}^{L}\sum_{i \in |D|} -(P^{l}_{y^{c}_{i}} \log(1-P^{l}_{\widehat{y}^{c}_{i}})+(1-P^{l}_{y^{c}_{i}}) \log P^{l}_{\widehat{y}^{c}_{i}}  )
  \label{eqn16}
\end{equation}

\begin{equation}
  Loss_{pair}=\sum_{l = 1}^{L}\sum_{i,j \in |D|\times|D|} -(P^{l}_{y^{pair}_{i,j}} \log(1-P^{l}_{\widehat{y}^{pair}_{i,j}})+(1-P^{l}_{y^{pair}_{i,j}}) \log P^{l}_{\widehat{y}^{pair}_{i,j}}  )
  \label{eqn16}
\end{equation}

In addition, 
we propose sort loss measure the performance that 
the $h^{'}$ exceeds $h$ as follows:

\begin{equation}
  Loss_{sort}=\sum_{l = 2}^{L} max(0,P^{l-1}_{y^{pair}_{i,j}=1}-P^{l}_{y^{pair}_{i,j}=1})+0.05
  \label{eqn16}
\end{equation}

where $P_{y^{pair}_{i,j}=1}$ is the prediction
of the labeled clause pair (ECP), $P^{l-1}$ is the output 
from $h^{l-1}$, and $P^{l}$ is the output 
from $h^{l}$. 
$Loss_{sort}$ ensures that $P^{l}$ is more effective 
than $P^{l-1}$, 
and ultimately ensures that $A^{(l)}$ is more effective 
than $A^{(l-1)}$. 

Note that $Activation Sort$ has no gradient as shown in Figure~\ref{figsort}, 
so $Loss_{sort}$ only measures the parameters after 
$Activation Sort$ in forward, and the parameters 
of the previous Enhanced GAT module not be affected. 
Due to the characteristics of $Loss_{sort}$, 
which increases $P^{l}_{y^{pair}}$ 
and also inhibits the improvement of $P^{l-1}_{y^{pair}}$. 
It is gradient stopping 
that can prevent the deviate gradient descent
of the $Loss_{sort}$ from conflicting 
with the other three losses. 

\section{Experiments}\label{ex}

Extensive experiments are conducted to verify the 
generality and effectiveness of our model 
on the Chinese ECPE dataset proposed by~\citet{xia-ding-2019-emotion} 
and the English RECCON dataset (ECPE-like task) 
proposed by~\citet{poria2021recognizing}. 
And we use the same metrics named $F1$ score for evaluating Chinese corpus following \citet{xia-ding-2019-emotion} 
and $avgF1$ score (the average of $F1$ scores for both positive and negative examples) for evaluating English corpus following \citet{poria2021recognizing}. 
\subsection{Dataset}

\begin{table}[hbp]
    \footnotesize
    \centering
    \begin{tabular}{|c|c|c|c|c|}
      \hline
      \textbf{dataset}&\textbf{document} & \textbf{sentence} & \textbf{clause} &\textbf{ECP}\\
      \hline
      Chinese&1945&8075&28727&2156\\
      English&1106&5513&11104&5380\\
      \hline
    \end{tabular}
    \caption{ Statistics of two datasets, where ECP stands for the 
    emotion-cause pair.}
    \label{tab2}
  \end{table}

  \begin{figure}[hbp]
    \centerline{\includegraphics[width=0.5\linewidth]{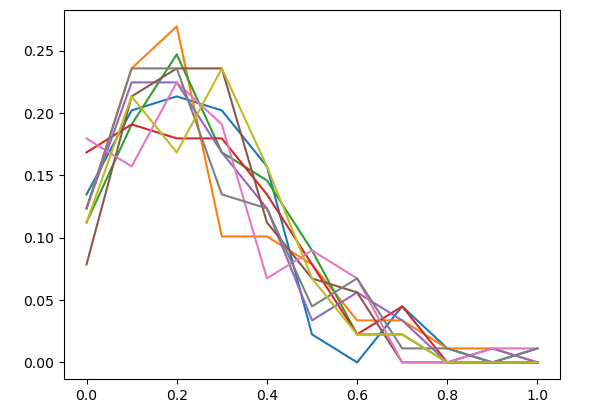}}
    \caption{Frequency diagram of sampling in contiguous 10 
    index of documents, where horizontal axis denotes the rate of 
    sample numbers of 10 documents group, vertical axis stands for the 
    correspond frequency.}
    \label{figfre}
  \end{figure}
  
  \begin{table*}[hbp]
    \footnotesize
    \centering
    \resizebox{\linewidth}{!}{
    \begin{tabular}{|c|c|c|c|c|c|c|c|c|c|c|}
      \hline
      \textbf{Statistic}&\textbf{fold1}&\textbf{fold2}&\textbf{fold3}&\textbf{fold4}&\textbf{fold5}&\textbf{fold6}&\textbf{fold7}&\textbf{fold8}&\textbf{fold9}&\textbf{fold10}\\
      \hline
      mean&342.78& 323.11& 353.30& 338.88& 340.09& 369.71& 326.13& 321.85& 336.10&338.4\\
      std&257.62& 246.80& 257.40& 251.27& 258.12& 246.92& 248.42& 239.52& 236.03&239.5\\
      skew ($\times10^{4}$)&1.71&2.17&1.66&2.39&2.07&1.14&2.45&2.21&2.04&2.11\\
      kurt ($\times10^{5}$)&-1.49&-1.49&-1.49&-1.49&-1.49&-1.49&-1.49&-1.49&-1.49&-1.49\\
      \hline
      std (Figure.1)&0.086& 0.086& 0.084& 0.077& 0.082& 0.091& 0.080& 0.083& 0.086&0.086\\
      \hline
  \end{tabular}}
  \caption{ Statistic in random 10-fold, we summarize the mean value, 
  standard deviation, skewness and kurtosis of the test dataset document-ids. Additionally, 
  we add the standard deviation of the sample frequencty. }
  \label{tab10fold}
  \end{table*}

  ~\citet{xia-ding-2019-emotion} constructs the Chinese dataset 
based on the ECE corpus~\citep{gui-etal-2016-event} and 
publishes the word segmentation version without periods. But 
to help reproduction and discussions, we refer to the 
ECE corpus and complete the ``.'', ``?'' and ``!'' 
to state the end of a sentence. 

While the Chinese ECPE corpus is a well-known and pioneering dataset 
for the ECPE task, it increasingly reveals some limitations that 
complicate the current works. Sparsity is the most concerning 
point. Statistically, in ECPE corpus, a document involves 14.32 clauses, 
3.56 sentences but 1.11 labeled pairs. And following the 
original splits, id-continuous samples are grouped together, when 
many long and complete documents are divided into 2-5 id-continuous samples. 
Hence, the sample variance is too numerically large and the 
feature distribution is substantially different between the 
train split and test split. Under the random initialization, 
the trivial solution exists, leading to an unstable learning process. 
Some works conditionally avoid this by deleting several samples from 
train split, while we adopt an alternative strategy that more makes sense. 
We initial the learnable parameters of our model with a pre-trained 
model from few-shot learning with 800 samples in train split. 

Besides, the English 
dataset is about dyadic conversations. So we 
define all utterances from the same speaker 
belong to a sentence. The details of two datasets 
are shown in Table\ref{tab2}. It has the 
almost 17 times less ratio of non-ECPs to ECPs than ECPE datasets. 
So although it is a neonate in the exploitation, it is more potential 
for relationship discovery and construction. But the original split 
is 3-fold, and the test dataset uses a contiguous sample strategy 
making us concerned about whether the bias exists to noise the experiment. 
To this end, we conduct a random 10-fold split for our and subsequent 
research. Table.\ref{tab10fold} demonstrates that our split is stochastic 
in the document distribution and provides the crucial statistics. 
Figure,\ref{figfre} visualizes the sampling frequency in each 10-contiguous 
samples, which indicates that our sampling is also independent. 

In order to obtain statistically 
credible performance, we evaluate our approach 10 times with 
different data splits by separately following
~\citet{xia-ding-2019-emotion,poria2021recognizing}. Additionally, 
we perform one sample $t-$test with $p$<0.01 to ensure the 
significance. 

\subsection{Implementation Details}

We replace the clause-level encoding part of 
all existing approaches with our model, 
with reservation of the other strategies and hyperparameters 
of corresponding existing approaches, besides   
the dimensions of E-GAT set to 768.  
For instance, when we embed our model into RankCP, we keep the layer of GAT with 2, 
learning rate with 0.001, weight decay with 1e-5, and the RBF kernel with $\sigma_{K}=1$. 
As the same as the word embedding of corresponding dataset baseline, 
we use $BERT_{Chinese}$ pre-trained model
\footnote{\url{http://github.com/huggingface/pytorch-pretrained-BERT}}, 
word vectors pre-trained from Chinese Weibo
\footnote{\url{http://www.aihuang.org/p/challenge.html}} for the experiment without BERT 
in the Chinese dataset, and $RoBERTa_{base}$ pre-trained model 
\footnote{\url{https://huggingface.co/roberta-base}}  
in the English dataset. 

And our entire codes and dataset splits are shared 
\footnote{\url{https://drive.google.com/file/d/1S_l59YYZfOXiS1HAFz39Y3g9GdqtCqGI/view?usp=sharing}}. 
Because the initial checkpoint models take up too much storage space. We will 
publish them to Github after the final version. 

\subsection{Baselines}

We compare our model with almost all published works, which can be 
divided into two categories: one is standard baselines following the 
benchmark~\citet{xia-ding-2019-emotion}, the other one is designed some 
auxiliary annotation or proposed extra data samples. We detailed them 
as follows: 

\textbf{Inter-EC}. A two-step method proposed by~\citet{xia-ding-2019-emotion} 
is the first to extract emotions and causes separately. And then to form 
an ECP, it train a binary classifier to filter out ECPs and non-ECPs. 

\textbf{RankCP}. A GAT-based method~\citep{wei-etal-2020-effective} 
tackles ECP prediction from a ranking perspective, adopting a 
graph attention network to propagate information among clauses. 

\textbf{ECPE-2D}. A transformer-based method~\citep{ding-etal-2020-ecpe} to 
construct a pairs matrix and achieve the interaction between ECPs by 
leveraging a 2D transformer module. 

\textbf{TDGC}. A transition-based method~\citep{fan-etal-2020-transition} 
formulated the ECPE task as a set of actions and transitions with directed 
graph construction. 

\textbf{PairGCN}. A GCN-based method~\citep{chen-etal-2020-end} constructed 
the influence relationships among candidate pairs by regarding pairs 
as nodes. 

\textbf{SLSN}. A local search method~\citep{cheng-etal-2020-symmetric} 
extracted the local ECPs by two novel corss-subnetwork of symmetric 
subnetworks separately. 

\textbf{SAP}. A relative position method~\citep{huang2022deep} extracted 
ECPs by defining the spans of two clauses among all pairs to achieve  
better performance. 

\textbf{RSN}. A recurrent-based method~\citep{chen2022recurrent} performs 
multiple rounds of inference to recognize emotion clauses, cause clauses, 
and emotion-cause pairs iteratively. 

\textbf{JointNN}. A self-attention-based method~\citep{tang2020joint} 
adopted BERT as word-level encoder and proposed a joint attention 
model for pair-level encoding. 

\textbf{MLL}. A slide window method~\citep{ding-etal-2020-end} employs 
an emotion-pivoted cause extraction framework and a cause-pivoted emotion 
extraction framework. 

\textbf{CPAM}. A new benchmark~\citep{chen-etal-2020-conditional} determined 
whether or not an ECP has a causal relationship given some specific 
context clauses in a corresponding improved dataset via manual annotation. 

\textbf{IE-CNN}. A new benchmark~\citep{chen-etal-2020-unified} proposed 
a new tagging strategy including four causal identity labels and seven 
emotion type labels. 

\textbf{BERT+}. A BERT-based model~\citep{cao2022research} achieved 
better performance following the benchmark as CPAM. 

\textbf{CL-ECPE}. A new contrastive benchmark~\citep{zhang2022cl} 
used adversarial samples as new datasets to perform the pair extraction. 

\textbf{MGSAG}. A new benchmark~\citep{bao2022multi} incorporated 
fine-grained and coarse-grained semantic representation by extracting 
keywords. 

\subsection{Overall Results}\label{sec5.3}

\begin{table*}[h]
    \footnotesize
    \centering
    \resizebox{\linewidth}{!}{
    \begin{tabular}{|c|c|ccc|ccc|ccc|}
      \hline 
      \multirow{2}{*}{\textbf{Category}} & \multirow{2}{*}{\textbf{Model}} & \multicolumn{3}{c}{emotion extraction(\%)} & \multicolumn{3}{c}{cause extraction(\%)} & \multicolumn{3}{c}{pair extraction(\%)}\\
      \cline{3-11}
      & & P & R  & F1 &P  & R  & F1 &P  & R  & F1\\
      \hline
      \multirow{16}{*}{\thead{without\\BERT}} & Inter-EC & 82.63 & 81.32 & 81.92 & 69.24 & 61.04 & 64.75 & 66.11 & 56.22 & 60.64\\
                             &+Ours & \textbf{82.89} & \textbf{81.94} & \textbf{82.41$\pm$1.75} & \textbf{71.33} & \textbf{65.07} &\textbf{68.06$\pm$2.41} & \textbf{68.47} & \textbf{59.31} & \textbf{63.56$\pm$2.11}\\
  \cline{2-11}
                                    & RankCP        & 87.03 & 84.06 & 85.48 & 69.27 & 67.43 & 68.24 & 66.98 & 65.46 & 66.10\\
                            & +Ours   & \textbf{88.16} & \textbf{85.87} &\textbf{87.00$\pm$1.65} &\textbf{72.82} & \textcolor{red}{\textbf{74.35}} &\textcolor{red}{\textbf{73.58$\pm$2.94}} &\textbf{69.54} & \textcolor{red}{\textbf{68.91}} & \textbf{69.22$\pm$2.53}\\
  \cline{2-11}
                                    & ECPE-2D  & 85.12 &\textbf{82.20} & 83.58 & \textbf{72.72} & \textbf{62.98} & \textbf{67.38} & 69.60 & \textbf{61.18} & 64.96\\
                            & +Ours  & \textbf{86.33} & 81.15 & \textbf{83.66$\pm$1.14} & 72.69 & 62.54 & 67.23$\pm$1.58 & \textbf{70.36} & 61.05 & \textbf{65.38$\pm$1.67}\\
  \cline{2-11}
                                    & TDGC     & 80.80 & 84.39 & 82.56 & 67.42 & 65.34 & 66.36 & 65.15 & 63.54 & 64.34\\
                            & +Ours     & \textbf{81.55} & \textbf{84.93} &\textbf{83.21$\pm$1.73} & \textbf{68.91} & \textbf{69.55} & \textbf{69.22$\pm$2.12} & \textbf{68.02} & \textbf{65.21} & \textbf{66.59$\pm$2.09}\\
  \cline{2-11}
                                    & PairGCN  & 85.87 & 72.08 & 78.29 & 72.83 & 59.53 & 65.41 & 69.99 & 57.79 & 63.21\\
                            & +Ours  & \textbf{86.39} & \textbf{73.31} & \textbf{79.31$\pm$1.66} & \textbf{76.41} & \textbf{64.28} & \textbf{69.82$\pm$2.27} & \textbf{71.06} & \textbf{62.41} & \textbf{66.45$\pm$1.95}\\
  \cline{2-11}
                                    & SLSN     & 84.06 & 79.80 & 81.81 & 69.92 & \textbf{65.88} & 67.78 & 68.36 & 62.91 & 65.45\\
                            & +Ours    & \textbf{84.74} & \textbf{80.69} & \textbf{82.67$\pm$1.33} & \textbf{73.53} & 64.21 &\textbf{68.55$\pm$2.03} & \textbf{70.00} & \textbf{63.68} & \textbf{66.69$\pm$1.79}\\
  \cline{2-11}
                                    & SAP   & \textbf{86.31} & \textbf{81.58} & \textbf{83.83} & 70.11 & \textbf{64.42} & 67.09 & \textbf{72.18} & 58.92 & 64.75\\
                                    & +Ours   & 85.59 & 79.12 & 82.23$\pm$1.63 & \textbf{72.39} & 64.29 & \textbf{68.10$\pm$2.59} & 71.56 & \textbf{62.93} & \textbf{66.97$\pm$2.31}\\
  \cline{2-11}
                                    &RSN       &86.88  & 87.43& 87.07& 73.62 & 65.54& 69.26& \textcolor{red}{72.15}& 63.77& 67.62\\
                &+Ours       & \textcolor{red}{\textbf{89.72}} & \textcolor{red}{\textbf{88.95}} & \textcolor{red}{\textbf{89.33$\pm$2.13}} & \textcolor{red}{\textbf{73.86}} & \textbf{69.46} & \textbf{71.59$\pm$2.56} & \textbf{71.55} & \textbf{67.42} & \textcolor{red}{\textbf{69.42$\pm$2.25}} \\
  \cline{2-11}
      \hline
      \multirow{14}{*}{\thead{with\\BERT}}    & JointNN  & \textbf{89.90} & 80.00 & 84.70 & $-$   & $-$   & $-$   & 71.10 & 60.70 & 65.50\\
                             & +Ours  & 88.25 & \textbf{84.56} & \textbf{86.37$\pm$2.51} & $-$   & $-$   & $-$   & \textbf{74.58} & \textbf{66.21} &\textbf{70.15$\pm$3.06}\\
  \cline{2-11}	
                            & RankCP        & 91.23          & 89.99 & 90.57 & 74.61 & 77.83 & 76.15 & 71.19 & 76.30 & 73.60\\
                            & +Ours   & \textcolor{red}{\textbf{92.70}} & \textbf{91.67} &\textcolor{red}{\textbf{92.16$\pm$2.82}} &\textbf{77.39} & \textcolor{red}{\textbf{81.12}} &\textcolor{red}{\textbf{79.18$\pm$1.45}} &\textbf{74.25} & \textcolor{red}{\textbf{79.79}} & \textcolor{red}{\textbf{76.88$\pm$1.46}}\\
  \cline{2-11}
                                    & ECPE-2D  & 86.27 & \textcolor{red}{\textbf{92.21}} & \textbf{89.10} & 73.36 & 69.34 & 71.23 & 72.92 & 65.44 & 68.89\\
                            & +Ours  & \textbf{88.59} & 89.12 & 88.85$\pm$3.59 & \textbf{74.55} & \textbf{69.82} & \textbf{72.11$\pm$2.01} & \textbf{73.62} & \textbf{66.08} & \textbf{69.65$\pm$2.37}\\
  \cline{2-11}
                          
                                    & TDGC     & 87.16 & 82.44 & 84.74 & \textbf{75.62} & 64.71 & 69.74 & \textbf{73.74} & 63.07 & 67.99\\
                            & +Ours     & \textbf{88.09} & \textbf{84.51} & \textbf{86.26$\pm$2.15} & 74.99 & \textbf{69.92} & \textbf{72.37$\pm$1.94} & 73.45 & \textbf{67.40} & \textbf{70.29$\pm$1.83}\\
  \cline{2-11}
                            & PairGCN  & 88.57 &79.58 & 83.75 & 79.07 & 68.28 & 73.27 & 76.92 & 67.91 & 72.02\\
                            & +Ours  & \textbf{90.55} & \textbf{82.9} & \textbf{86.56$\pm$2.62} & \textcolor{red}{\textbf{79.13}} & \textbf{71.83} & \textbf{75.30$\pm$1.89} & \textcolor{red}{\textbf{78.41}} & \textbf{71.09} & \textbf{74.57$\pm$2.34}\\
  \cline{2-11}
                            &MLL& 86.08 & \textbf{91.91} &\textbf{88.86}  & 73.82 & \textbf{79.12} &76.30 & 77.00 & 72.35 &74.52 \\
                            &+Ours &\textbf{89.64} & 86.13 & 87.85$\pm$1.31 &\textbf{78.34} & 75.89 & \textbf{77.10$\pm$1.03} &\textbf{77.62} & \textbf{73.11} & \textbf{75.29$\pm$1.79} \\
  \cline{2-11}
                &RSN       &86.14  & 89.22& 87.55& 77.27 & 73.98& 75.45& 76.01& 72.19& 73.93\\
                &+Ours       & \textbf{87.11} & \textbf{90.41} & \textbf{88.73$\pm$1.96} & \textbf{77.81} & \textbf{75.63} & \textbf{76.70$\pm$2.27} & \textbf{76.84} & \textbf{75.30} & \textbf{76.06$\pm$2.10} \\
      
      \hline
    \end{tabular}}
    \caption{Performance comparison in ECPE Chinese corpus for replacing the 
    clause-level encoding of all existing approaches with our model (+Ours). 
    Except for RANKCP with GAT and JointNN with BiLSTM+attention 
    for clause-level encoding, the other approaches above 
    adopt BiLSTM to model the clause representation. 
    Results significantly show that 
    our model is more effective than 
    original networks of all existing approaches. }
    \label{tab3}
  \end{table*} 
  
  \begin{table}
    \footnotesize
    \centering
    \resizebox{0.5\linewidth}{!}{
    \begin{tabular}{|c|c|c|c|}
      \hline
      \multirow{2}{*}{model}&\multicolumn{3}{c}{Scores (\%)}\\
      \cline{2-4}
       &P&R&F1\\
      \hline
      CPAM&\textbf{61.95}&78.65&69.29\\
      +Ours&60.72$\pm$1.51&\textbf{84.33$\pm$2.35}&\textbf{72.60$\pm$1.94}\\
      \hline
      IE-CNN&81.86&64.96&66.86\\
      +Ours&\textbf{82.44$\pm$3.81}&\textbf{68.92$\pm$2.27}&\textbf{67.41$\pm$2.56}\\
      \hline
      BERT+&65.03&82.07&72.54\\
      +Ours&\textbf{66.41$\pm$1.24}&\textbf{84.67$\pm$3.34}&\textbf{75.41$\pm$2.14}\\
      \hline
      CL-ECPE$\dagger$&$\diagdown$&$\diagdown$&81.95\\
      +Ours&$\diagdown$&$\diagdown$&\textbf{82.21$\pm$0.79}\\
      \hline
      MGSAG&\textbf{87.17}&77.12&75.21\\
      +Ours&86.82$\pm$1.89&\textbf{81.53$\pm$2.05}&\textbf{76.39$\pm$1.58}\\
      \hline
    \end{tabular}}
    \caption{ The statistics of 5 methods with auxiliary annotation, 
    extra data or samples.}
    \label{tabaux}
  \end{table}
  
  \begin{table*}
    \footnotesize
    \centering
    \resizebox{\linewidth}{!}{
    \begin{tabular}{|c|c|ccc|ccc|ccc|}
      \hline 
      \multirow{2}{*}{\textbf{Category}} & \multirow{2}{*}{\textbf{Model}} & \multicolumn{3}{c}{emotion extraction(\%)} & \multicolumn{3}{c}{cause extraction(\%)} & \multicolumn{3}{c}{pair extraction(F1(\%))}\\
      \cline{3-11}
      & & P  & R  & F1 &P  & R  & F1 &pos  & neg  & avg\\
      \hline
      \multirow{4}{*}{\thead{with\\RoBERTa}}    & RankCP & \textbf{90.70} & 62.48 & 73.96$\pm$2.98 & \textbf{79.62} & 57.73 & 66.92$\pm$1.44 & 51.45 & 97.06 & 74.26$\pm$1.13\\
      &+Ours & 90.08 & \textbf{64.41} & \textcolor{red}{\textbf{75.08$\pm$3.82}} & 79.00 & \textbf{60.76} & \textbf{67.07$\pm$1.45} & \textbf{52.56} & \textbf{97.14} & \textbf{74.85$\pm$1.46}\\
  \cline{2-11}
                         & ECPE-2D  & 72.17 &41.33 & 52.56$\pm$1.92 & 81.44 & 61.49 & 54.83$\pm$3.65 & 51.07 & 97.11 & 74.09$\pm$2.02\\
     & +Ours  & \textbf{88.57} &\textbf{59.11} & \textbf{70.90$\pm$1.29} & \textbf{82.59} & \textbf{63.81} & \textcolor{red}{\textbf{71.99$\pm$4.35}} & \textbf{53.75} & \textbf{97.37} & \textcolor{red}{\textbf{75.56$\pm$1.87}}\\
      \hline
    \end{tabular}}
    \caption{Performance comparison in RECCON English corpus, 
    due to lacking word vectors pre-trained with word2vec toolkit, 
    we cancel the experiment without RoBERTa.}
    \label{tab4}
  \end{table*} 

  As shown in Table~\ref{tab3}, 
for evaluating the generality 
and effectiveness of our model in ECPE dataset, we embed our model to 8 
existing approaches without BERT and 7 approaches with BERT 
to summarize the performances on three tasks 
(emotion extraction, cause extraction, pair extraction). 
In approaches without BERT, the word2vec is adopted 
in word embedding, while the pre-training BERT model fine-tuned 
is adopted for approaches with BERT. 

Overall, our model generally outperforms the traditional clause-level 
encoding model (BiLSTM or GAT) in all published approaches. Note that 
our model does not adopt any plug-in technology, so these results 
demonstrate that our clause-level encoding model is better than all 
existing clause-level encoding model. 
Besides, our model significantly improves the performance 
of graph-based neural network baselines 
(i.e., 
RankCP with $3.12\%$ raise without BERT and $3.28\%$ with BERT, 
PairGCN with $3.24\%$ raise without BERT and $2.55\%$ with BERT), which 
indicates that our model leverages the unique advantages GNNs enjoy 
such as the use of part neighboring nodes for aggregation. 
And in the graph-agnostic works, 
TDGC with $2.25\%$ and $2.3\%$ improvement, Inter-EC with $2.92\%$, 
JointNN with $4.65\%$, RSN with $1.8\%$ and $2.13\%$ improvement. 
These works entail a more straightforward pair-level encoding 
than those with complex downstream models 
such as a hybrid encoding for both clause-level and pair-level. 
Although our model advances less 
in these hybrid encoding works due to the part of superposition 
in clause relationship learning, this result reassures that clause 
representation orienting to clause-to-clause relationship defined by 
ours is effective. 

Moreover, to further evaluate the performance in other language 
or text genres corpus, as shown in Table\ref{tab4} 
we conduct the experiments on the English conversation dataset
named RECCON for the causal emotion entailment task that is 
same as ECPE task. Due to the different text genres, 
the results in the RECCON are a little 
different in increments but demonstrate that the relationship 
proposed by ours made our model broadly applicable.  

To further indicate the generality of our clause-level encoding method, 
we tested the performance in other benchmarks in Table~\ref{tabaux} 
by replacing their clause-level encoding model with ours. 
From Table~\ref{tabaux}, CPAM+Ours and BERT++Ours perform better than 
other benchmarks under the unique conditional annotation dataset their 
enjoy. Hence, tagging the causal relationship and adding negative samples 
benefit our model to model the clause-to-clause relationship. And CL-ECPE 
improves least among all benchmarks. We infer that the adversarial samples 
introduce more confounders in clause-to-clause relationship. The same 
situation occurs in IE-CNN benchmark which introduces more 
relation-agnostic lable types. 

\subsection{Ablation Study}

\begin{table}
    \footnotesize
    \centering
    \resizebox{0.5\linewidth}{!}{
    \begin{tabular}{|c|c|c|c|}
      \hline
      \textbf{task}&\textbf{Model} & \textbf{cn: F1(\%)} & \textbf{en: avgF1(\%)} \\
      \hline
      \multirow{5}{*}{C} & Ours & 79.18 &53.1\\
      &-$Loss_{sort}$ &-0.19 &-1.81\\
      &-$Sort$&-0.88 &-3.26\\
      &-$Loss_{sort}$-$Sort$&-1.11&-4.91\\
      &-E-GAT &\textbf{-2.66}& \textbf{-6.44}\\
      \hline
      \multirow{5}{*}{E} & Ours &92.16 &60.92\\
      &-$Loss_{sort}$ &-0.65&-0.63\\
      &-$Sort$&-0.88&-1.12 \\
      &-$Loss_{sort}$-$Sort$&-1&-1.18\\
      &-E-GAT &\textbf{-1.51}&\textbf{-7.07} \\
      \hline
      \multirow{5}{*}{P} & Ours & 76.88&67.33\\
      &-$Loss_{sort}$ &-0.6&-0.86\\
      &-$Sort$&-0.81&-1.33\\
      &-$Loss_{sort}$-$Sort$&-1.07&-1.05\\
      &-E-GAT &\textbf{-2.73}&\textbf{-2.14}\\
      \hline
    \end{tabular}}
    \caption{ Ablation analysis on RANKCP embedded with EA-GAT. 
    To facilitate discussion, we show the experiment in 
    ECPE corpus with BERT and experiment in RECCON dataset 
    without BERT why both have the most obvious gap in the 
    4 ablation settings. C, E, and P stand for cause extraction, 
    emotion extraction, and pair extraction. Then cn and en denote the 
    Chinese corpus and English corpus. }
    \label{tab5}
  \end{table}

  To explore the contribution of E-GAT, Activation Sort,  
and $Loss_{sort}$,
3 sets of ablation experiments with the same baseline 
on 2 datasets are conducted. In Table\ref{tab5}, we investigate  
results under the following cases: removing the $Loss_{sort}$ 
function; replacing Activation Sort with a $\tanh$ function; 
both above cases; and replacing E-GAT with GAT.   
One observes that replacing E-GAT leads to the 
highest degradation in performance while modifying 
$Loss_{sort}$ and Activation Sort yields fewer losses. 
And this situation occurs in the two subtasks of 
RECCON corpus more sharply. 
There are two 
reasons we surmise. One is that in the ECPE corpus, 
the number of sentences of each sample is as $2.075$ times 
as the number in RECCON, where the fewer sentences yield a more 
favorable attention distribution. The other one is that 
the ratio of ECP to non-ECP in RECCON is $1:10.31$  
too much more than $1:174.44$ in ECPE 
to not need to widen the gap among 
the non-ECPs. In a word, they all indicate that 
clause (utterance) relationship is the crucial component that 
existing works lack, while Activation Sort is more like a 
strategy used for specific data distribution.

\subsection{Case Analysis}\label{sec5.5}

\begin{figure}[hbp]
    \centerline{\includegraphics[width=0.5\linewidth]{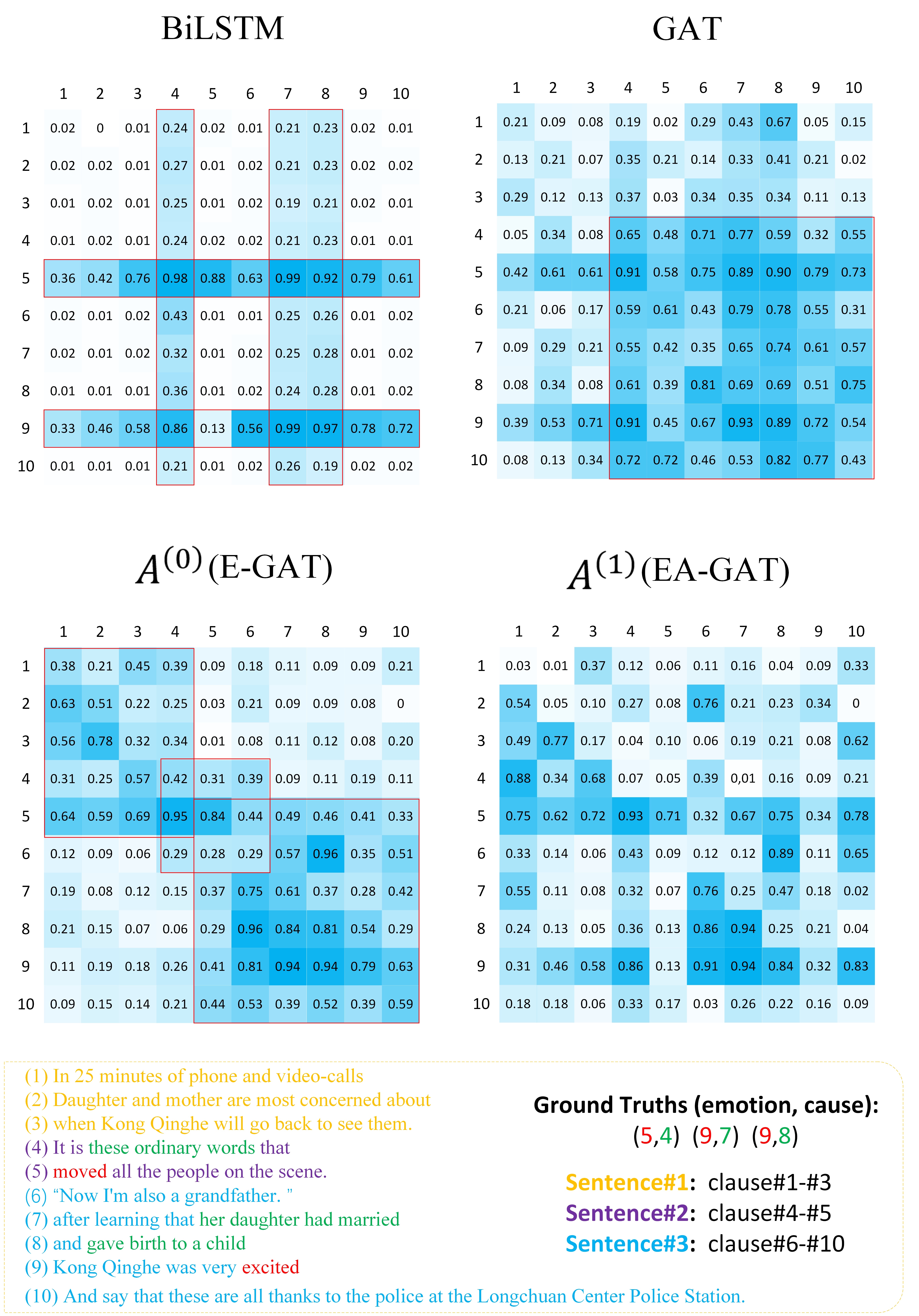}}
    \caption{Adjacency matrices of BiLSTM, GAT,  $A^{(0)}$ and $A^{(1)}$}. 
    \label{figcase}
  \end{figure}

  To further illustrate how our model improves the performance, 
  we inspect the matrix computed from BiLSTM, GAT, 
  E-GAT ($A^{(0)}$), and 
  EA-GAT ($A^{(1)}$) of the test dataset 
  given by the ``RankCP+Ours with BERT'' model in ECPE dataset 
  with one visualization shown in Figure\ref{figcase}. 
  
  In BiLSTM, all non-ECPs have almost zero weight except 
  the row where the emotion clause is located and the column 
  where the cause clause is located.  
  It indicates that BiLSTM 
  fails to identify whether pairs such as (5,7) (5,8) 
  are  correct ECPs. 
  
  In GAT, the relationship weights between non-ECPs 
  are different, but  
  the weight distribution is concentrated near 
  the emotion/cause clauses, and the non-ECPs 
  far away from the ECPs still lack effective weight evaluation.  
  
  In $A^{(0)}$ (E-GAT), 
  what is different from the above two is, the weight distribution 
  is not concentrated near the specific row and column or area, 
  but is markedly distinguished to 3 parts via 3 sentences. 
  And unlike the above two, the weights of (5,7) and (5,8) 
  are significantly lower than the correct ECP such as (5, 4), 
  this is because weights of $(5,6)$, $(6,7)$, and $(7,8)$ 
  are not high, which indicates that there 
  is no reverse causal influence from $8$-th clause 
  to $5$-th clause.
  
  In $A^{(1)}$ (EA-GAT), it is obvious that 
  there is a series of stepped weight distributions, 
  such as $(2,1) (3,2) (4,3) (5,4)$, which matches the fact 
  that most clauses are the cause of 
  the next clause in the document.  
  This result corroborates the theory of
  ~\citet{mann1988rhetorical,marcu2000theory}, 
  who state that when the clause relationship is aligned with the 
  reasoning process 
  (in this case, multi-mask attention and activation sort), 
  the model 
  learns to exploit underlying semantic structure more easily.

\subsection{Sensitivity Aanlysis}

In this section, we further investigate how the different activation 
tricks and the layer $l$ would affect the performance. In addition, 
we also analyze the sensitivity in between the long document and 
short document. 

\begin{figure}[hbp]
    \centerline{\includegraphics[width=0.5\linewidth]{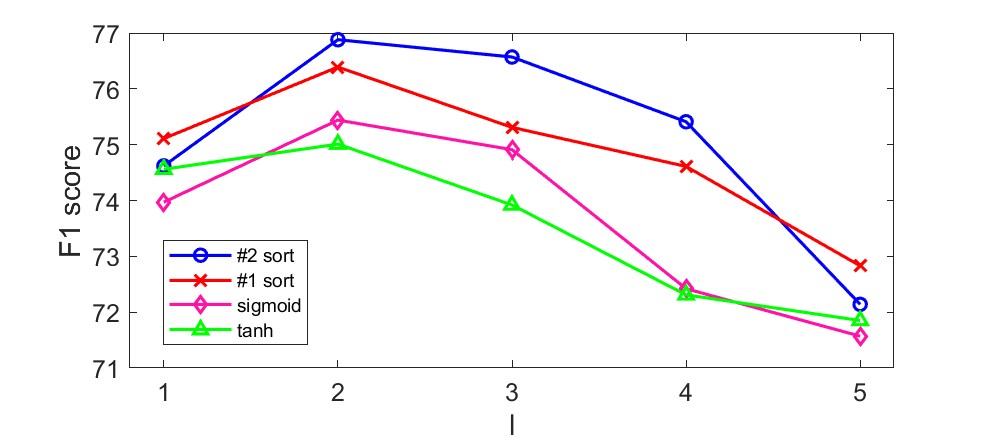}}
  \caption{The results of different layers of EGAT and activation tricks 
  under the baseline of RankCP with BERT }. 
  \label{figlayer}
\end{figure}

\begin{table}[hbp]
    \footnotesize
    \centering
    \begin{tabular}{|c|c|c|c|c|}
      \hline
     \textbf{category} & \textbf{Approach} & P($\%$) & R($\%$) & F1($\%$) \\
     \hline
     \multirow{2}{*}{$<$15} & RANKCP & 70.96 & 76.33 & 73.55\\
      &Ours& 74.68 & 79.81 & \textbf{77.15} \\
  
      \hline
      \multirow{2}{*}{$>=$15} & RANKCP & 72.38 & 74.93 & 73.63 \\
      & Ours& 73.94 & 79.76 &\textbf{ 76.72} \\
     \hline
  \end{tabular}
    \caption{ Comparative results for documents with 2 sets of clause numbers}
    \label{tablong}
  \end{table}

In Figure~\ref{figlayer}, we conducted a set of contrasts with the 
layers of EGAT ($l$) up to 4 and four activation strick: $\#$2 sort 
following Equation~\ref{eqn12}; $\#$1 sort following Equation~\ref{eqn11}; 
\textit{Sigmoid} activation function; and \textit{tanh} activation function. 
From Figure~\ref{figlayer}, there are three observation: (i) Activation Sort($\#$2 sort and $\#$1 sort) 
both achieved better performance than activation functions ($sigmoid$ and $tanh$), 
which collaborate the standpoint in Section~\ref{subas} that popular activation 
functions can not wipe the distance among non-ECPs. (ii) $\#$2 sort achieved 
the best performance, which indicate that Equation~\ref{eqn12} is better 
designed to highlight the weights of ECPs. (iii) $l=2$ is sufficient 
to yeild the most effective clause representation while $l=1$ is not due 
to lacking of Activation Sort.

And to analyze the performance of EA-GAT to length of documents, 
we divided the entire dataset into approximately equal 2 parts as shown in Table~\ref{tablong}, 
one for documents with less than 15 clauses 
and the other one with more than or equal to 15 clauses.  
Then RANKCP is used as the baseline for comparison. 
In the category more than 15, RANKCP shows better performance, 
while our model increases $3.09\%$ but shows less performance 
comparatively to the short documents. 
It demonstrates that the EA-GAT network is more suitable for documents 
with a short number of clauses. 

\section{Results and Discussion}\label{rd}

Results are shown in Table\ref{tab3},\ref{tab4},\ref{tab5} and Figure\ref{fig3}, 
and we have following observations:

\begin{itemize}
  \item \textbf{EA-GAT is effective. } 
  Both under word embedding with BERT and word2vec (without BERT), 
  the $F1$ score of pair extraction significantly 
  improve by $2.35\%$ and $1.89\%$ averagely. 
  Moreover, the $F1$ score of cause extraction improve by 
  $1.77\%$ and $2.75\%$ respectively in BERT and without BERT and 
  the $F1$ score of emotion extraction improve by $1.17\%$ and $0.89\%$. 
  The most improvement occurs in cause extraction which indicates 
  that traditional graph-based or recurrence-based model is 
  insufficient for clause-to-clause representation learning. And it is 
  interesting to see that cause extraction is enhanced more 
  in word2vec embedding and emotion extraction more in BERT embedding. 
  We consider that emotion extraction depends on word embedding more. 
  But they both demonstrate that our model can 
  achieve better predictive power by strengthening one of or both subtasks. 
  \item \textbf{EA-GAT is general. } 
  We detail our comprehensive empirical evaluation from 3 different 
  angles to support the generality of our model: (i) our model is integrated with 10 standalone baselines 
  with an averagely improve $2.14\%$ in pair extraction 
  and $0.98\%$, $2.65\%$ in emotion/cause extraction separately; 
  (ii) both in Chinese narrative corpus and 
  English conversation corpus, our model has significant advantage 
  with $2.1\%$ and $1.03\%$ on average respectively; (iii) from 
  the 3 different upstream word embedding methods (word2vec, BERT, and RoBERTa), 
  our model substantially surpasses all the baselines with 
  $1.89\%$, $2.35\%$, and $1.03\%$ average improvement respectively. 
\end{itemize}

While the EA-GAT is generally applicable in ECPE task, there are 
some limitations of EA-GAT.

Activation Sort is a radical framework/strategy. We define 
it as redical, not bacause it stop the gradient, but because it 
maps a clause pair into fixed distribution. Institutionally, 
stoping the gradient of a intermediate result absolutely 
destroy the whole back propagation. But practically, 
as the one of the inputs of the second E-GAT module, it is 
affected by two types of loss functions. One is the task loss 
including emotion prediction loss, cause prediction loss and 
pair prediction loss which measure the predictive power. According 
to the input $A^{(0)}$, the word encoding part can learn a 
favorable $A^{(0)}$ representation for prediction losses. 
Identically, $\widehat{y}^{e}$ and $\widehat{y}^{e}$ is the 
input of the second E-GAT also the output of the first E-GAT, 
which could lead to that $\widehat{y}^{e}$ and $\widehat{y}^{e}$ 
lose the purity of thier own prediction results of subtasks. 
Such a unnecessarty confilct more substantially occur in sort loss. 
Hence, if we used the traditional activation function (i.e., RelU or tanh) 
to replace the Sort mapping, we also had to make 
$\widehat{y}^{e}$ and $\widehat{y}^{e}$ lose gradients. 

What we think that it is imperfect is, for each document, we define 
a invariant sequence order for mapping though there are some 
alternative sequences in Section 4.3. In this case, each document 
has a fixed clause relationship level that should not be the same. 
In other words, this is our most concerned problem: how to 
construct a clause relationship? The clause relationship, a 
unique structure to each document, indicate that which clauses 
are irrelevant to the observed clause and which are closely related. 
Like a priori knowledge for a neural network to learn the unique 
structure among clauses rather than a full-linked structure that 
learn a clause representation from all the other clauses. For this 
perspective, the clause-level relationship defined by ours is a 
breakthrough which fine the grain of clause relationship from the 
whole document to sentence. And we attempt to advance the grain 
to the single clause via Sort mapping, while the performance of 
Sort is not as satisfying as clause-level relationship. 

\section{Conclusion}\label{co}

In this paper, we have defined the new and systematic clause-level 
relationship for the ECPE task. And to exploit this relationship, 
we have developed EA-GAT, a GAT-based model 
with multi-mask attention and activation mapping. It produces 
an effective clause representation by aggregating the information 
from different clause-to-clause relationships and incorporating the 
sort position embedding of subtask prediction sequences to achieve grain consistency. 
With comprehensive experiments in the two emotion-cause datasets named ECPE and RECCON,   
the generality and effectiveness of our model are demonstrated. 
Furthermore, we conduct the ablation studies and case analysis to 
show that clause-level encoding is an integral contributor to the ECPE task.




\bibliographystyle{elsarticle-num-names} 
\bibliography{extra}





\end{document}